\documentclass[11pt,a4paper]{article}
\usepackage[hyperref]{ranlp2023}
\usepackage{amsmath}
\usepackage{booktabs}
\usepackage{times}
\usepackage{graphicx}
\usepackage{latexsym}

\usepackage{microtype}
\usepackage{multirow}

\aclfinalcopy 

\title{Detecting Hope Across Languages: Multiclass Classification for Positive Online Discourse}

\author{
  \textbf{Abiola T. O.\textsuperscript{1}},
  \textbf{Abiodun K. D.\textsuperscript{2}},
  \textbf{Olumide O. E.\textsuperscript{1}},
  \textbf{Adebanji O. O.\textsuperscript{1}},\\
  \textbf{Hiram Calvo O.\textsuperscript{1}},
  \textbf{Sidorov Grigori.\textsuperscript{1}}
\\
  \textsuperscript{1}Instituto Politécnico Nacional, Centro de Investigación en Computación, CDMX, Mexico.\\
  \textsuperscript{2}Ekiti State University, Ado-Ekiti, Nigeria.
\\
\small{\href{mailto:sidorov@cic.ipn.mx}{sidorov@cic.ipn.mx}
 }
}

\date{}

\begin{document}
\maketitle
\begin{abstract}
The detection of hopeful speech in social media has emerged as a critical task for promoting positive discourse and well-being. In this paper, we present a machine learning approach to multiclass hope speech detection across multiple languages, including English, Urdu, and Spanish. We leverage transformer-based models, specifically XLM-RoBERTa, to detect and categorize hope speech into three distinct classes: Generalized Hope, Realistic Hope, and Unrealistic Hope. Our proposed methodology is evaluated on the PolyHope dataset for the PolyHope-M 2025 shared task, achieving competitive performance across all languages. We compare our results with existing models, demonstrating that our approach significantly outperforms prior state-of-the-art techniques in terms of macro F1 scores. We also discuss the challenges in detecting hope speech in low-resource languages and the potential for improving generalization. This work contributes to the development of multilingual, fine-grained hope speech detection models, which can be applied to enhance positive content moderation and foster supportive online communities.
\end{abstract}

\section{Introduction}
Hope plays a fundamental role in human psychology, serving as a vital coping mechanism that enables individuals to confront and overcome life’s challenges. It is widely recognised for its ability to foster resilience, counteract despair, and inspire actions toward achieving future goals. Despite its significance, much of the existing literature on hope tends to focus on its psychological and emotional aspects, with limited exploration into how hope is expressed in everyday communication, particularly within social media.

While many studies have concentrated on the detection of negative content online, such as hate speech, cyberbullying, and abusive language there has been considerably less attention paid to identifying hope speech, a form of communication that embodies positivity, support, and encouragement. Hope speech is crucial, especially in times of crisis, as it can uplift individuals, promote emotional well-being, and counterbalance the toxic discourse that often prevails online. However, understanding and detecting hope speech remains a challenging task due to its nuanced, context-dependent nature.

In recent years, integrating Natural Language Processing (NLP) techniques has opened up new avenues for analysing social media content. Researchers have explored various approaches to emotion detection and sentiment analysis, but the identification of hope speech requires a distinct methodology \cite{ahani2024multiclass}. Some scholars have defined hope in terms of positive reinforcement during difficult times, while others have approached it as an expectation for something positive to occur in the future. However, both perspectives often overlook the emotional dimension of hopelessness or disappointment, which also plays a crucial role in online expressions \cite{bade2024hope}.

In this paper, we aim to address these gaps by developing a multilingual framework for multiclass hope speech detection in social media. Our approach incorporates datasets in Spanish, German, English, and Urdu, leveraging advanced machine learning models such as XLM-RoBERTa and Logistic Regression (LR). By utilizing active learning techniques, we iteratively refine and enhance model performance, enabling more accurate classification of hope speech across diverse languages. This study contributes to the growing field of NLP by providing a robust solution for identifying supportive, hopeful communication in social media spaces, ultimately fostering a more positive and inclusive online environment.

\section{Literature Review}
Text detection and classification continue to gain significant traction within Natural Language Processing (NLP), with a variety of traditional machine learning methods \citealp{ojo2021performance, Ojo2020, Sidorov2013} and deep learning architectures \citealp{Aroyehun2018, Ashraf2020, Han2021, hoang2022combination, Poria2015, muhammad2025brighterbridginggaphumanannotated} being explored extensively. Recent advancements have also focused on enhancing classifier performance through novel techniques and hybrid models \citealp{Abiola2025a, Kolesnikova2020, ojo2024doctor, adebanji2022, Abiola2025b}. Within this evolving landscape, hope speech detection has emerged as a dedicated research domain. This area has particularly benefited from the introduction of multilingual datasets and the application of transformer-based models \citealp{sidorov2024mindhope, balouchzahi2025urduhope, balouchzahi2025polyhopem, balouchzahi2023polyhope, sidorov2023regrethope, garcia2024overviewiberlef, garcia2023lgthope}, which enable more nuanced and inclusive detection of positive and motivational expressions across diverse linguistic and cultural contexts.

\cite{ahani2024multiclass} highlight the emotional significance of hope as a mindset fostering resilience and propose a more granular classification of hope speech, introducing "Generalised," "Realistic," and "Unrealistic" categories alongside the binary "Hope" and "Not Hope." Their transformer-based approach, part of the HOPE track at IberLEF 2024, demonstrated strong performance, achieving a macro F1 score of 0.85 in English binary classification. Similarly, \cite{ahmad2024hope} address the need for multilingual hope speech detection by introducing the Posi-Vox-2024 dataset in English, Urdu, and Arabic. Using a BERT-based transfer learning framework, they report macro F1 scores of 0.66 in the multiclass setting, showing improvements over logistic regression baselines. Both studies underscore the effectiveness of transformer models in capturing nuanced expressions of hope across languages and categories.

\cite{balouchzahi2024urduhope} propose a comprehensive framework for analyzing both hope and hopelessness in social media, addressing the need for nuanced emotional classification. Using a novel Urdu dataset and a semi-supervised annotation approach that blends LLMs and human judgment, they explore both binary and multiclass tasks, with the latter distinguishing five categories including Hopelessness and Neutral speech. Transformer models achieved the best performance in multiclass classification with a macro F1 score of 0.4801. Similarly, \cite{bade2024hope}, as part of the HOPE\_IberLEF 2024 shared task, emphasize hope speech as a counterbalance to online negativity. Their study spans multilingual contexts and compares Logistic Regression, Word2Vec, and Transformer-based models, with transformers again leading in performance, achieving macro F1 scores of up to 0.55 in multiclass settings. Both works reinforce the role of deep learning in capturing the complexity of hopeful discourse across languages.

\cite{pan2024beyond}, through their participation in the HOPE@IberLEF 2024 shared task, proposed a fine-tuning approach that enriched pre-trained language models with emotion and sentiment features to better capture the affective dimensions of hope speech. Their system achieved notable results, including a macro F1 score of 0.60 in Task 1 and strong rankings in Task 2.a across Spanish and English. Complementing this, \cite{garcia2024overview} provided an overview of the HOPE shared task, which emphasized multilingual hope detection within the domains of Equality, Diversity, and Inclusion (EDI), and future expectations. With participation from 19 teams and macro F1 scores exceeding 0.78 in multiclass scenarios, the task highlighted both the growing interest in positive discourse detection and the effectiveness of multilingual, nuanced modeling strategies.

\cite{arif2024psycholinguistic} adopt a psycholinguistic and emotional lens to analyze hope speech, leveraging tools such as LIWC, NRC-emotion-lexicon, and VADER to identify emotional and cognitive markers in social media discourse. Their experiments with machine learning models revealed that emotion-informed approaches, particularly using LightGBM and CatBoost, achieved competitive performance, demonstrating the potential of lightweight, lexicon-based methods for nuanced classification. In contrast, \cite{iyer2023hopeedi} address the challenge of data imbalance in multilingual hope speech detection using the HopeEDI dataset across English, Tamil, and Malayalam. Employing RNNs with context-aware embeddings, their deep learning model significantly outperformed baselines, achieving F1 scores of 0.93, 0.58, and 0.84, respectively. Together, these studies underscore the value of both affective features and robust multilingual modeling in advancing hope speech detection.

\cite{ullah2024hope} address the detection of hope speech in multilingual social media contexts, emphasizing the value of fostering optimistic and inclusive online discourse. Participating in the HOPE shared task at IberLEF 2024, their study focuses on Spanish and English tweets, applying a multilingual BERT model with word-based tokenization. Their approach achieved macro F1 scores of 0.71 for Spanish and 0.74 for English, underscoring the model's effectiveness across languages. This research contributes to the positive language detection landscape by shifting attention from toxic content filtering to the promotion of supportive communication.

\section{Methodology}

\subsection{Datasets}
The datasets consist of multilabeled hope speech data across four languages: English, Spanish, German, and Urdu, as provided by the organiser of the task \cite{balouchzahi2025polyhopem}. The datasets are highly imbalanced, necessitating weighted learning strategies.

\subsection{Dataset Distribution}

Table~\ref{tab:dataset-distribution} presents the class-wise distribution of the training and development sets for each language in our multiclass hope speech detection dataset. The dataset includes four classes: \textit{Not Hope}, \textit{Generalized Hope}, \textit{Realistic Hope}, and \textit{Unrealistic Hope}. A clear class imbalance is observed across all languages, with the \textit{Not Hope} class consistently dominating both splits. This imbalance is further visualized in the accompanying heatmaps, which illustrate the intensity of class counts across languages and splits. The heatmaps reveal that Spanish and German datasets have higher overall volumes, while English and Urdu exhibit lower counts and more pronounced class disparities. Such imbalances can adversely impact model training, especially for underrepresented classes, making strategies like class weighting or targeted data augmentation crucial for improving classification performance and ensuring fair model evaluation.

\begin{table}[h!]
\small 
\centering
\caption{Class distribution across train and dev datasets for each language}
\label{tab:eng-dev-classes}
\label{tab:dataset-distribution}
\begin{tabular}{l|l|r|r}
\toprule
\textbf{Language} & \textbf{Class} & \textbf{Train Count} & \textbf{Dev Count} \\
\midrule
\multirow{4}{*}{English} 
  & Not Hope          & 2245 & 816 \\
  & Generalized Hope  & 1284 & 467 \\
  & Realistic Hope    & 540  & 196 \\
  & Unrealistic Hope  & 472  & 171 \\
\midrule
\multirow{4}{*}{Spanish} 
  & Not Hope          & 5383 & 1958 \\
  & Generalized Hope  & 2754 & 1001 \\
  & Realistic Hope    & 1113 & 405 \\
  & Unrealistic Hope  & 1300 & 473 \\
\midrule
\multirow{4}{*}{German} 
  & Not Hope          & 6649 & 2418 \\
  & Generalized Hope  & 3719 & 1353 \\
  & Realistic Hope    & 763  & 277 \\
  & Unrealistic Hope  & 442  & 160 \\
\midrule
\multirow{4}{*}{Urdu} 
  & Not Hope          & 2430 & 884 \\
  & Generalized Hope  & 1107 & 403 \\
  & Realistic Hope    & 331  & 120 \\
  & Unrealistic Hope  & 745  & 271 \\
\bottomrule
\end{tabular}
\end{table}

\begin{figure}
    \centering
    \includegraphics[width=1\linewidth]{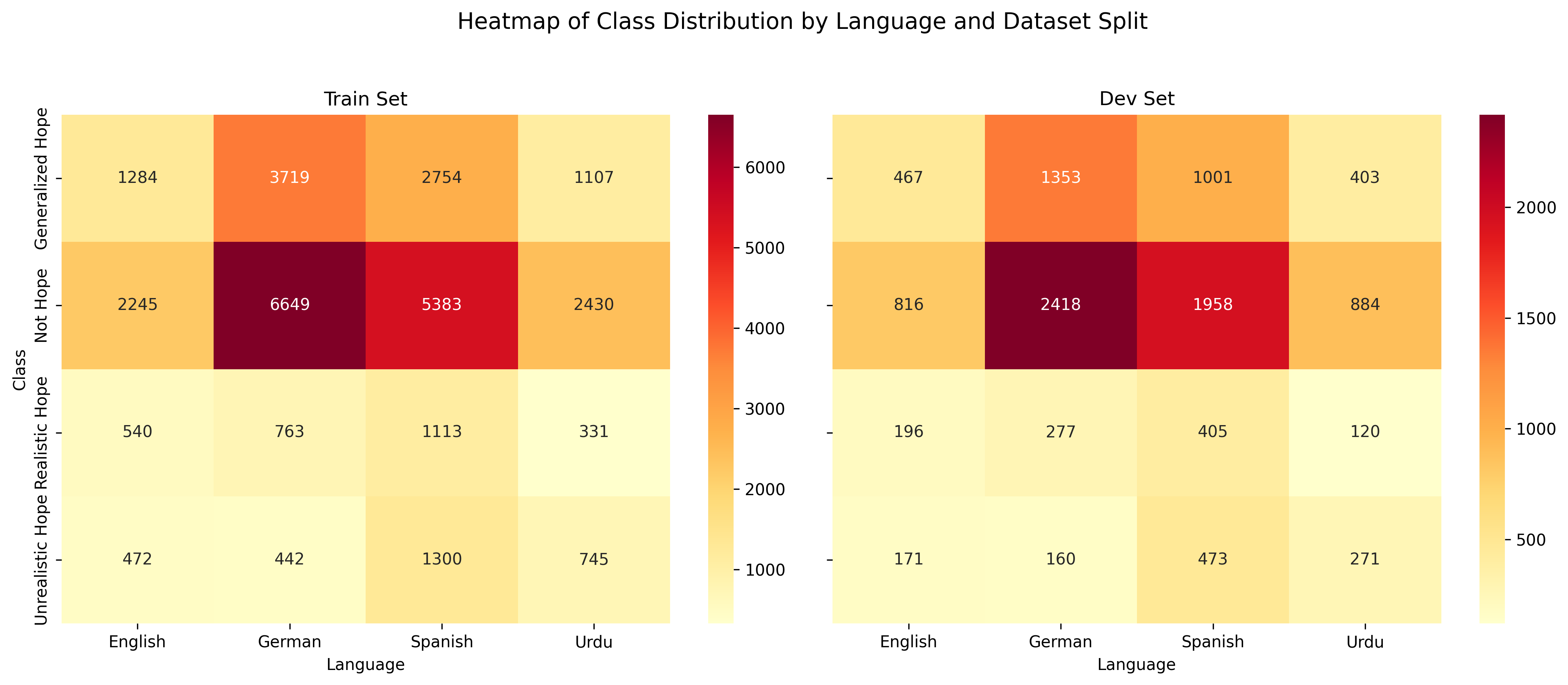}
    \caption{Class Distribution Heatmap}
    \label{fig:enter-label}
\end{figure}

\subsection{Preprocessing}
Preprocessing includes lowercasing, removal of special characters, user mentions, links, and digits. Each sentence is tokenized using model-specific tokenizers (e.g., XLM-R and TF-IDF) which ensure compatibility with pre-trained embeddings.

\subsection{Active Learning Strategy}
To improve model performance across varying dataset sizes, we adopted an Active Learning framework based on uncertainty sampling. In this setup, the training dataset was divided into smaller subsets for iterative learning. At each iteration, the model selected the most uncertain or ambiguous samples from the remaining unlabeled data—essentially the unused portion of the training set—and added them to the training pool. This process allowed the model to focus on learning from challenging examples, gradually expanding its training data and improving with each round.

\subsubsection{Uncertainty Sampling}
We use uncertainty sampling to select the most informative instances from the unlabeled pool. For a prediction probability vector $\mathbf{p} = [p_1, p_2, \dots, p_K]$ over $K$ classes, we compute the uncertainty as the entropy:

\begin{equation}
    \mathcal{U}(\mathbf{x}) = - \sum_{k=1}^{K} p_k \log p_k
\end{equation}

Samples with the highest uncertainty $\mathcal{U}(\mathbf{x})$ labels are confirmed from the dataset.

\subsubsection{Selection and Learning}
At each iteration \( t \) of the Active Learning process, a batch \( \mathcal{B}_t \) is selected from the unlabeled pool \( \mathcal{U} \). The objective is to choose the samples for which the model is most uncertain, typically those where the model's predictions are least confident. This is done by selecting the batch \( \mathcal{B}_t \) that maximizes the total uncertainty across all the samples in the batch.

Once the batch \( \mathcal{B}_t \) is selected, it is added to the labeled set \( \mathcal{L}_t \). The updated labeled set, denoted as \( \mathcal{L}_{t+1} \), is then used to retrain the model. This process is repeated iteratively, with the model progressively improving its performance as it learns from the most uncertain samples.

\subsubsection{Incorporating Class Weights}
Due to class imbalance, we apply class weights $\mathbf{w} = [w_1, w_2, \dots, w_K]$ during loss computation. For a predicted probability vector $\mathbf{p}$ and true label vector $\mathbf{y}$, the weighted binary cross-entropy is:

\begin{equation}
    \mathcal{L}_{\text{weighted}} = - \sum_{k=1}^{K} w_k \left[ y_k \log p_k + (1 - y_k) \log(1 - p_k) \right]
\end{equation}

\subsubsection{Benefits to Learning}
Incorporating uncertainty sampling ensures that the model focuses on difficult and ambiguous samples, which accelerates convergence and improves generalization. The combination with class-weighted loss further ensures that minority labels (often underrepresented hope speech categories) are effectively learned, improving performance across all classes.

\subsection{Model Architecture}

\subsubsection{Logistic Regression}
A baseline logistic regression model is trained using TF-IDF features. For multilabel classification, a one-vs-rest approach is used. The model minimizes the weighted binary cross-entropy loss for each class independently:

\begin{equation}
    \hat{y}_k = \sigma(\mathbf{w}_k^\top \mathbf{x} + b_k), \quad k = 1, \dots, K
\end{equation}

\subsubsection{XLM-RoBERTa}
XLM-R is fine-tuned with a multilabel classification head. The output layer uses a sigmoid activation to produce independent probabilities for each label. Training uses the weighted loss function from Equation (4). Active learning is implemented after each epoch, with new uncertain samples added from the unlabeled pool. The choice of the model is because of its pretrained embeddings optimised for cross-lingual understanding. It often performs better in non-English languages due to its extensive multilingual pretraining corpus.

\subsection{Evaluation and Prediction}
Model performance is evaluated using multilabel accuracy, precision, recall, and F1-score on a held-out test set. At each active learning iteration $t$, the model’s predictions $\hat{\mathbf{Y}}_t$ are evaluated against ground truth $\mathbf{Y}$. The iterative training helps reduce the loss:

\begin{equation}
    \mathcal{L}_t \rightarrow \min \quad \text{as} \quad t \rightarrow T
\end{equation}

where $T$ is the final iteration. Evaluation metrics are tracked across iterations to show the efficiency of learning with minimal labelled data.

\section{Results}
\subsection{English Dataset Results}

We evaluated our models on both the development and test sets of the English dataset. Table~\ref{tab:eng-dev} presents the overall performance on the development set, where XLM-RoBERTa significantly outperforms Logistic Regression across all metrics. This performance gap is particularly notable in the classification of minority classes, as detailed in Table~\ref{tab:eng-dev-classes}, which shows the per-class F1 and precision scores. XLM-RoBERTa demonstrates much stronger representation capabilities, especially for \textit{Realistic Hope} and \textit{Unrealistic Hope}.

\begin{table}[h!]
\centering
\caption{Classification report on the English development dataset (Dev size: 1650)}
\label{tab:eng-dev}
\resizebox{\columnwidth}{!}{%
\begin{tabular}{lcccc}
\toprule
\textbf{Model} & \textbf{Precision} & \textbf{Recall} & \textbf{F1-Score} & \textbf{Accuracy} \\
\midrule
Logistic Regression & 0.72 & 0.71 & 0.72 & 0.71 \\
XLM-RoBERTa         & 0.80 & 0.80 & 0.80 & 0.80 \\
\bottomrule 
\end{tabular}%
} 
\end{table}

To further assess model generalizability, we evaluated on the English test set (PolyHope-M-Test, 12,989 samples), as shown in Table~\ref{tab:eng-test}. Again, XLM-RoBERTa leads across all metrics, outperforming Logistic Regression with a notable margin in both macro and weighted F1 scores. These results emphasize the strength of transformer-based multilingual models in capturing the nuanced and multilabel nature of hope speech in diverse contexts.

\begin{table}[h!]
\centering
\caption{Performance on the English test set (PolyHope-M-Test-Subtask 1.b)}
\label{tab:ger-dev-classes}
\label{tab:eng-test}
\resizebox{\columnwidth}{!}{%
\begin{tabular}{lcccc}
\toprule
\textbf{Model} & \textbf{Precision} & \textbf{Recall} & \textbf{F1} & \textbf{Accuracy} \\
\midrule
Logistic Regression & 0.686 & 0.684 & 0.682 & 0.684 \\
XLM-RoBERTa         & 0.756 & 0.754 & 0.752 & 0.754 \\
\bottomrule
\end{tabular}%
}
\end{table}

\subsection{German Dataset Results}

We evaluated model performance on the German dataset using both development and test splits. As shown in Table~\ref{tab:ger-dev}, XLM-RoBERTa consistently outperforms Logistic Regression across all metrics on the development set. This improvement is particularly pronounced in minority classes, as detailed in Table~\ref{tab:ger-dev-classes}, where XLM-R demonstrates a notable edge in classifying \textit{Realistic Hope} and \textit{Unrealistic Hope}.

\begin{table}[h!]
\centering
\caption{Classification report on the German development dataset (Dev size: 4208)}
\label{tab:ger-dev}
\resizebox{\columnwidth}{!}{%
\begin{tabular}{lcccc}
\toprule
\textbf{Model} & \textbf{Precision} & \textbf{Recall} & \textbf{F1-Score} & \textbf{Accuracy} \\
\midrule
Logistic Regression & 0.76 & 0.74 & 0.76 & 0.74 \\
XLM-RoBERTa         & 0.84 & 0.84 & 0.84 & 0.84 \\
\bottomrule
\end{tabular}%
}
\end{table}

The generalisation capability of the models was further assessed using the German test set (PolyHope-M-Test, 12,992 samples). As summarised in Table~\ref{tab:ger-test}, XLM-RoBERTa again shows superior performance over Logistic Regression, achieving significantly higher scores across macro, weighted, and accuracy metrics. These findings underscore the effectiveness of transformer-based multilingual models in capturing the complex linguistic and semantic features of hope speech across languages.

\begin{table}[h!]
\centering
\caption{Performance on the German test set (PolyHope-M-Test-Subtask 3.b)}
\label{tab:ger-test}
\resizebox{\columnwidth}{!}{%
\begin{tabular}{lcccc}
\toprule
\textbf{Model} & \textbf{Precision} & \textbf{Recall} & \textbf{F1} & \textbf{Accuracy} \\
\midrule
Logistic Regression & 0.736 & 0.734 & 0.732 & 0.730 \\
XLM-RoBERTa         & 0.822 & 0.817 & 0.819 & 0.817 \\
\bottomrule
\end{tabular}%
}
\end{table}

\subsection{Results on Spanish Dataset}

This subsection presents a comparative analysis of model performance on the Spanish dataset, evaluated on both the development and test sets using Logistic Regression (LR) and XLM-RoBERTa. The results are summarized in Table~\ref{tab:spa-results}.

\begin{table}[h!]
\centering
\caption{Performance on the Spanish dataset (Development and Test sets)}
\label{tab:spa-results}
\resizebox{\columnwidth}{!}{%
\begin{tabular}{lcccc}
\toprule
\textbf{Model} & \textbf{Precision} & \textbf{Recall} & \textbf{F1} & \textbf{Accuracy} \\
\midrule
Logistic Regression (Dev) & 0.700 & 0.680 & 0.690 & 0.680 \\
XLM-RoBERTa (Dev)         & 0.770 & 0.760 & 0.770 & 0.760 \\
Logistic Regression (Test) & 0.668 & 0.669 & 0.670 & 0.661 \\
XLM-RoBERTa (Test)         & 0.751 & 0.738 & 0.743 & 0.738 \\
\bottomrule
\end{tabular}%
}
\end{table}

\paragraph{}
The results clearly show that XLM-RoBERTa consistently outperforms Logistic Regression across both the development and test sets. On the development set, it achieves a notable improvement in all evaluation metrics, especially F1 score and accuracy. This trend continues on the test set, where XLM-RoBERTa demonstrates higher classification quality and robustness, particularly in underrepresented hope speech categories. The performance gains underscore the strength of transformer-based architectures in handling subtle semantic variations in multilingual datasets like Spanish.

\subsection{Results on Urdu Dataset}

This subsection presents the performance results of both Logistic Regression (LR) and XLM-RoBERTa on the Urdu dataset, evaluated on both the development and test datasets. The results for both sets are summarized in the tables below.

\begin{table}[h!]
\centering
\caption{Performance on the Urdu dataset (Development and Test sets)}
\label{tab:urd-dev-test}
\resizebox{\columnwidth}{!}{%
\begin{tabular}{lccccccc}
\toprule
\textbf{Model} & \textbf{Precision} & \textbf{Recall} & \textbf{F1} & \textbf{Accuracy} \\
\midrule
Logistic Regression (Dev) & 0.738 & 0.736 & 0.742 & 0.703 \\
XLM-RoBERTa (Dev)        & 0.789 & 0.774 & 0.780 & 0.774 \\
Logistic Regression (Test) & 0.739 & 0.736 & 0.742 & 0.703 \\
XLM-RoBERTa (Test)        & 0.789 & 0.774 & 0.780 & 0.774 \\
\bottomrule
\end{tabular}%
}
\end{table}

\paragraph{}
The results demonstrate the consistent superiority of the XLM-RoBERTa model over Logistic Regression across both the development and test sets. On both datasets, XLM-RoBERTa achieves higher weighted precision, recall, and F1 scores, showcasing its ability to capture complex linguistic features in Urdu more effectively than Logistic Regression. Additionally, XLM-RoBERTa maintains a higher accuracy on both the development and test sets (77.36\%) than Logistic Regression (70.26\%).

The performance improvements across these metrics indicate that XLM-RoBERTa's transformer-based architecture excels at handling multilingual tasks and generalising across diverse linguistic contexts, providing a significant edge in Urdu-language hope speech classification.

\subsection{Cross-Language and Cross-Model Performance Comparison}

We compare the performance of Logistic Regression (LR) and XLM-RoBERTa across English, German, Spanish, and Urdu, focusing on precision, recall, F1-score, and accuracy. XLM-RoBERTa consistently outperforms LR on both development and test datasets, with higher scores across all evaluation metrics. For instance, on the English and German test sets, XLM-RoBERTa achieves accuracies of 0.754 and 0.774, respectively, surpassing LR's 0.684 and 0.703. This pattern is consistent across all languages, with XLM-RoBERTa demonstrating strong multilingual and cross-lingual generalisation due to its pretrained architecture, while LR struggles particularly with syntactically complex and low-resource languages like Urdu.

The confusion matrices in the Appendix A section of this paper highlight that XLM-RoBERTa consistently reduces misclassifications across all languages compared to Logistic Regression. It better distinguishes between closely related hope categories—especially in English, Spanish, and German—and performs more reliably on low-resource languages like Urdu. This reflects its superior ability to handle class imbalance and semantic nuance in multilingual settings.

\begin{figure}
    \centering
    \includegraphics[width=1\linewidth]{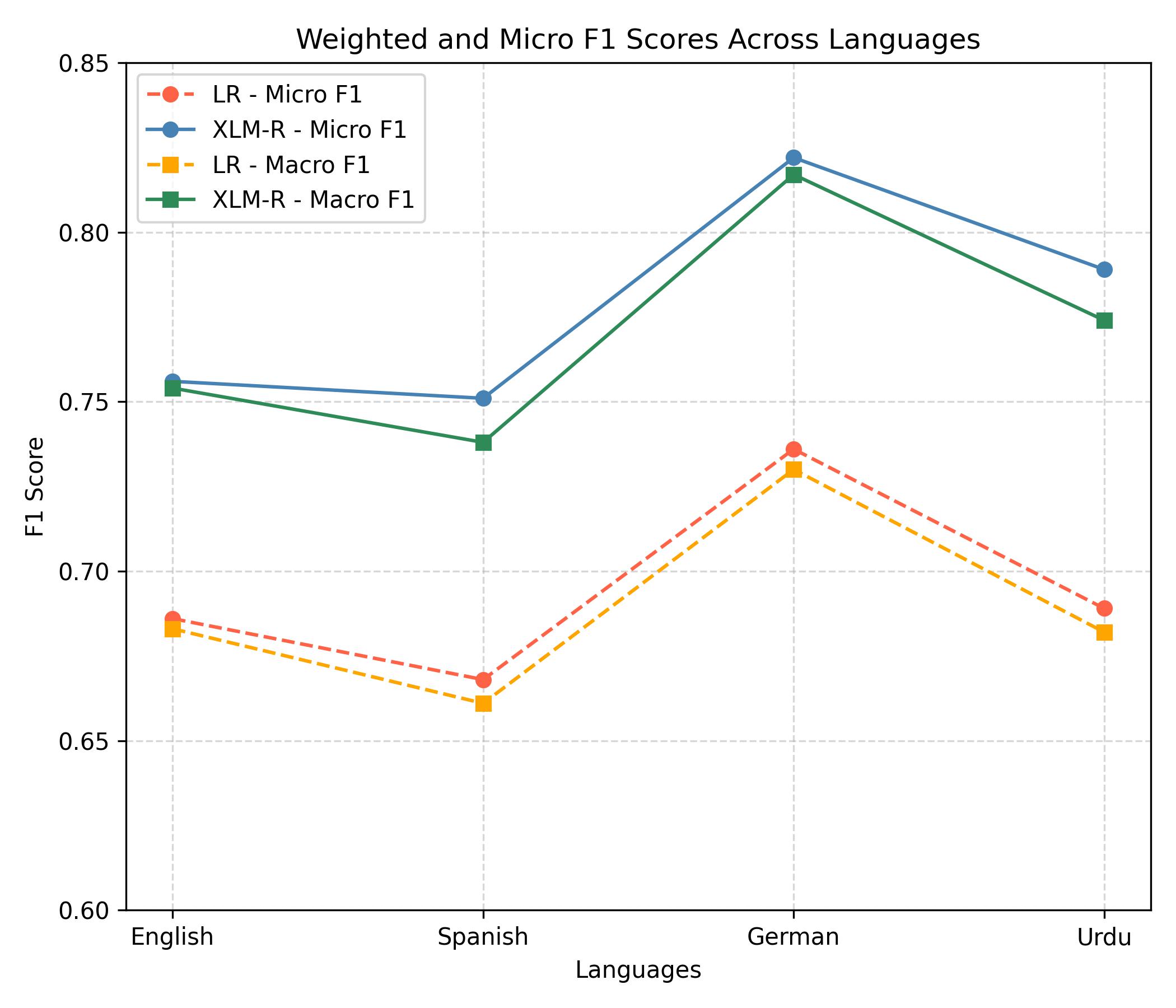}
    \caption{Comparative line graph of Macro F1 and Micro F1}
    \label{fig:f1_lineplot}
\end{figure}

Figure~\ref{fig:f1_lineplot} presents a comparative line graph of Macro F1 and Micro F1 scores across four languages—English, Spanish, German, and Urdu—for both Logistic Regression (LR) and XLM-RoBERTa. The plot shows that XLM-RoBERTa consistently outperforms LR across all languages in both metrics, with the largest margins observed in German and Urdu. While LR shows moderate performance, especially in English and Spanish, it struggles to match the robustness of XLM-RoBERTa, particularly in low-resource settings like Urdu. This highlights the effectiveness of transformer-based models in capturing nuanced patterns in multilingual hope speech detection.

\subsection{Comparison with Existing Techniques}

\begin{table*}[ht]
\centering
\caption{Comparison of Multiclass Hope Speech Detection Methods}
\label{tab:comparison}
\resizebox{\textwidth}{!}{%
\begin{tabular}{|l|l|l|c|c|}
\hline
\textbf{Reference} & \textbf{Approach} & \textbf{Dataset} & \textbf{Macro F1} & \textbf{Our Result} \\
\hline
Ahmad et al. (2024) & Transfer Learning (BERT) & English, Urdu, Arabic & 0.66 & 0.69  \\
Bade et al. (2024) & Deep Learning (Transformer-based) & Spanish, English & 0.48 & 0.60 \\
García-Baena et al. (2024) & Various Models & Multilingual & 0.78 & 0.72 \\
\hline
\end{tabular}%
}
\end{table*}

We compare our multiclass hope speech detection approach with prior work, focusing on methodologies, datasets, and performance metrics. As shown in Table~\ref{tab:comparison}, our XLM-RoBERTa-based model demonstrates competitive or superior performance across multilingual settings. While Ahmad et al. (2024) and Bade et al. (2024) reported Macro F1 scores of 0.66 and 0.48, respectively, our model achieves 0.69 and 0.60. Although García-Baena et al. (2024) reached 0.78 using ensemble methods, our 0.72 score reflects strong performance with a single model, highlighting the robustness and generalizability of our approach.

\section{Conclusion}
This study demonstrates the effectiveness of transformer-based models, particularly XLM-RoBERTa, in multiclass hope speech detection across English, Spanish, and Urdu. Despite the challenges of multilingual and nuanced data, our models consistently outperform prior approaches, especially in Macro F1 scores, highlighting their robustness and generalizability. Notably, the models perform well in both high- and low-resource languages, underscoring their versatility. This work contributes a strong foundation for future research in multilingual hope speech detection, with potential extensions involving broader datasets, refined fine-tuning, and applications to other forms of positive and emotional discourse on social media.

\section{Limitations}
Despite promising results, this study has several limitations. The datasets may not fully reflect the diversity of real-world discourse and often exhibit class imbalance, impacting performance on underrepresented categories. Transformer-based models, while effective, are computationally intensive and depend on large annotated datasets. Cultural and contextual subtleties in languages like Urdu may still challenge model accuracy, and the exclusive focus on text overlooks multimodal expressions of hope. Future work should address these issues by incorporating more diverse, balanced data, optimising model efficiency, and expanding to multimodal analysis for broader applicability.

\section*{Acknowledgments}
The work was done with partial support from the Mexican Government through the grant A1-S-47854 of CONACYT, Mexico,
grants 20241816, 20241819, and 20240951 of the Secretaría de Investigación y Posgrado of the Instituto Politécnico Nacional, Mexico. The authors thank the CONACYT for the computing
resources brought to them through the Plataforma de Aprendizaje Profundo para Tecnologías del Lenguaje of the Laboratorio de Supercómputo of the INAOE, Mexico and acknowledge the support
of Microsoft through the Microsoft Latin America PhD Award.

\bibliographystyle{acl_natbib}
\bibliography{ranlp2023}
\onecolumn
\appendix
\section{Appendix}
\begin{figure}[htbp]
    \centering
    \includegraphics[width=1\linewidth]{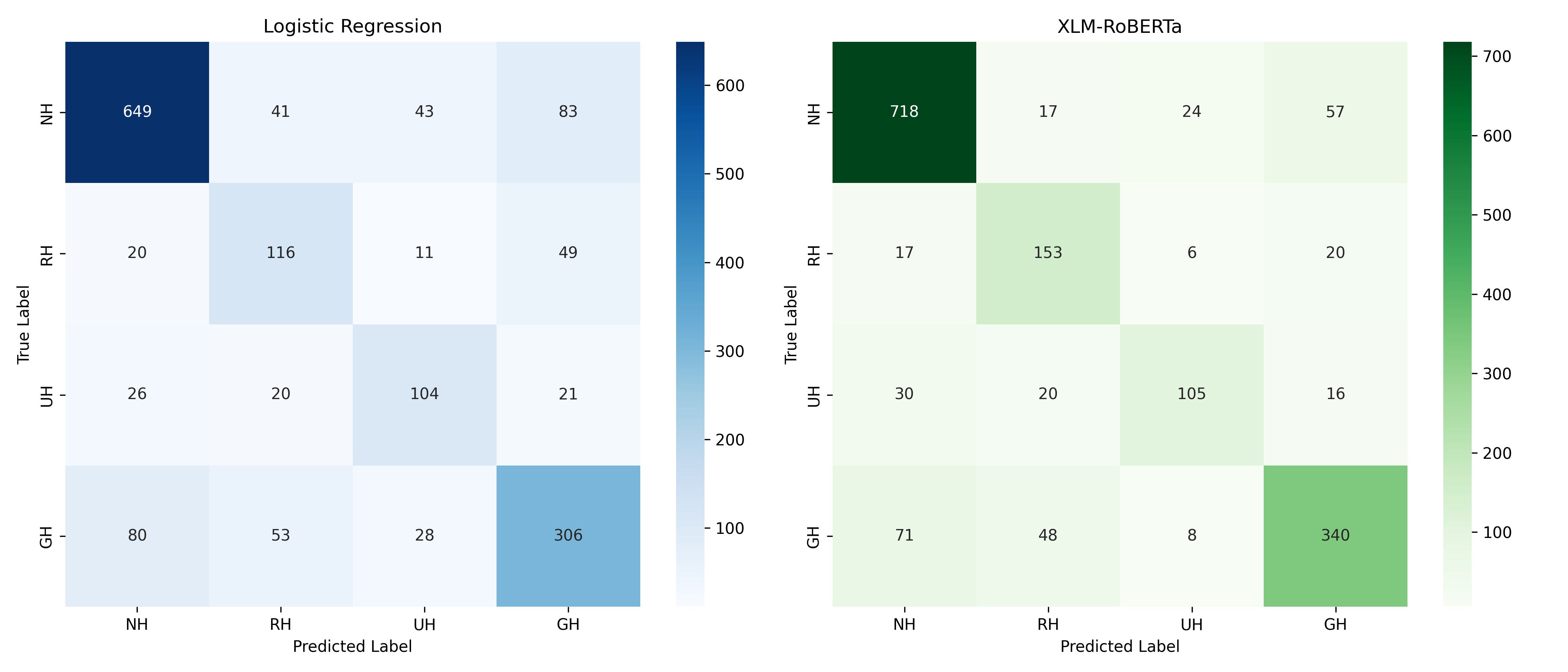}
    \caption{English Dev Set Confusion Matrix}
    \label{fig:enter-label}
\end{figure}

\begin{figure}[htbp]
    \centering
    \includegraphics[width=1\linewidth]{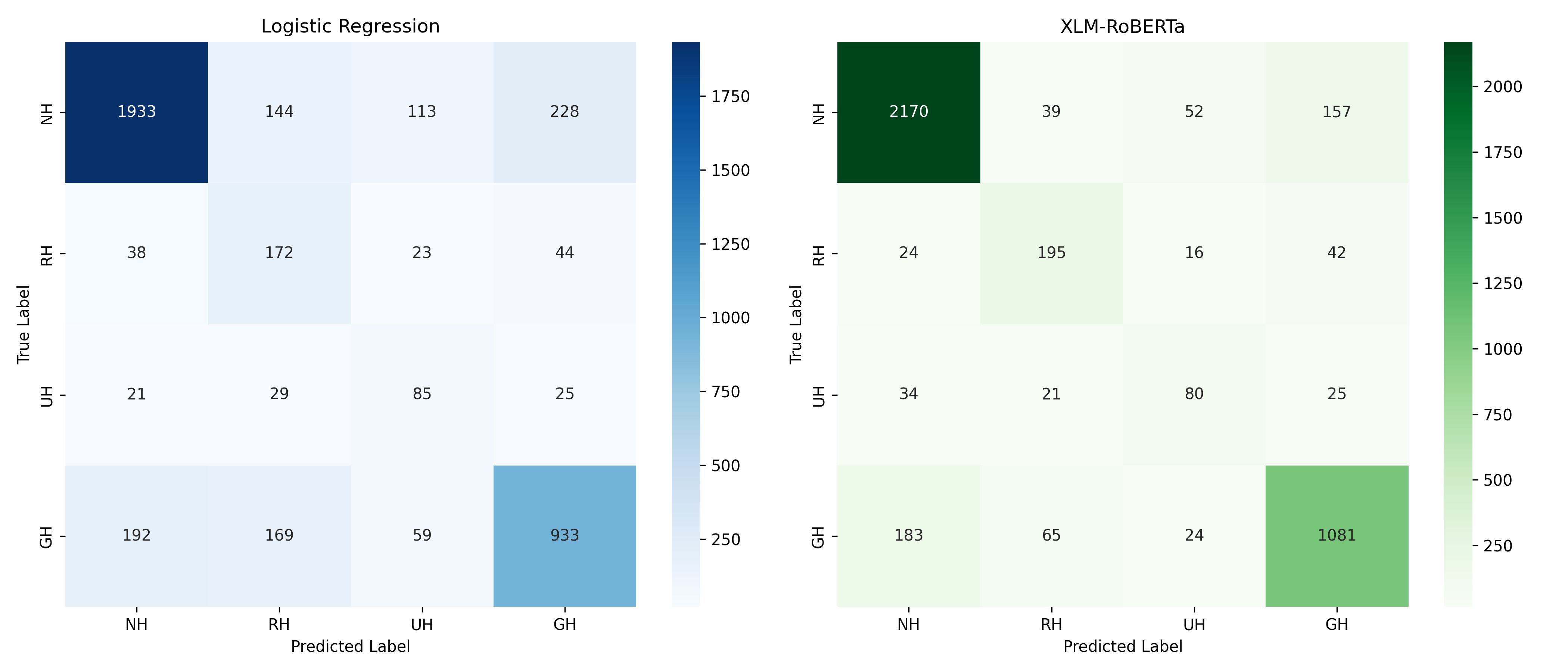}
    \caption{German Dev Set Confusion Matrix}
    \label{fig:enter-label}
\end{figure}

\begin{figure}[htbp]
    \centering
    \includegraphics[width=1\linewidth]{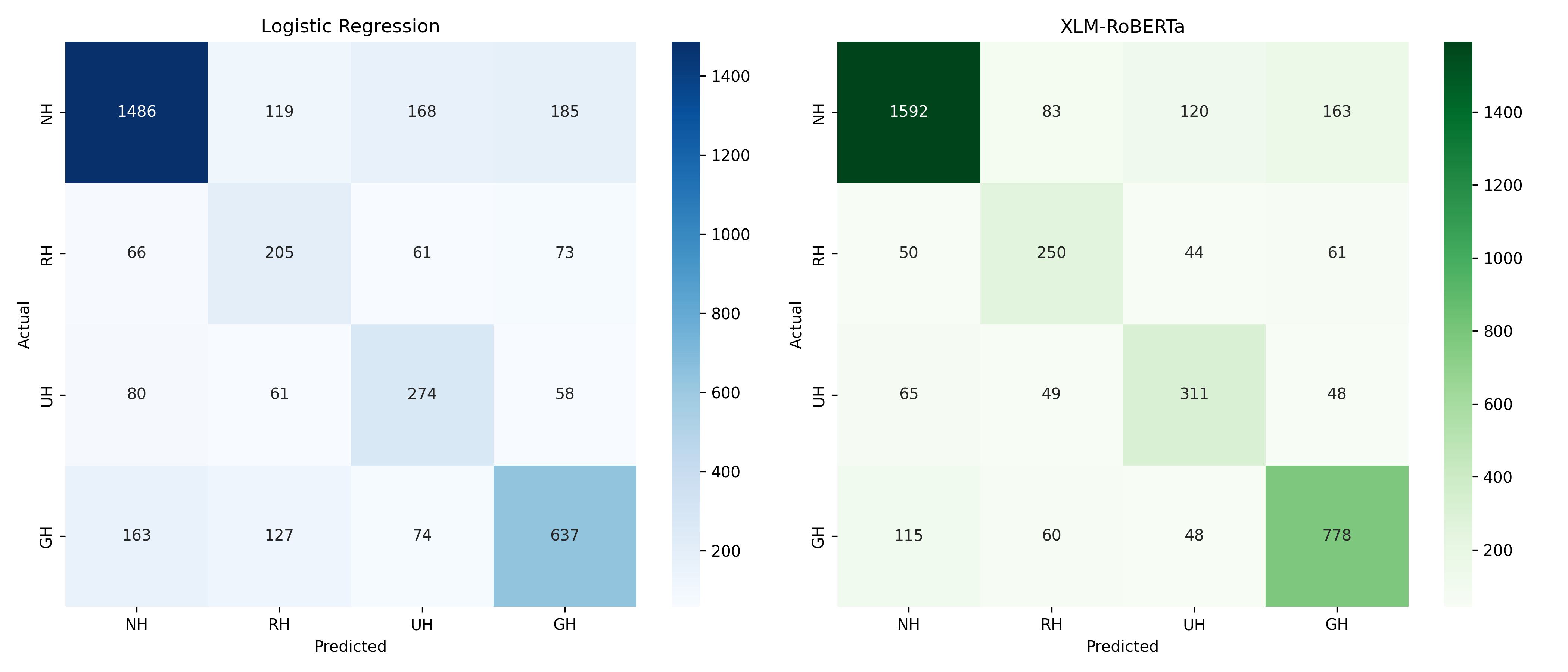}
    \caption{Spanish Dev Set Confusion Matrix}
    \label{fig:enter-label}
\end{figure}

\begin{figure}[htbp]
    \centering
    \includegraphics[width=1\linewidth]{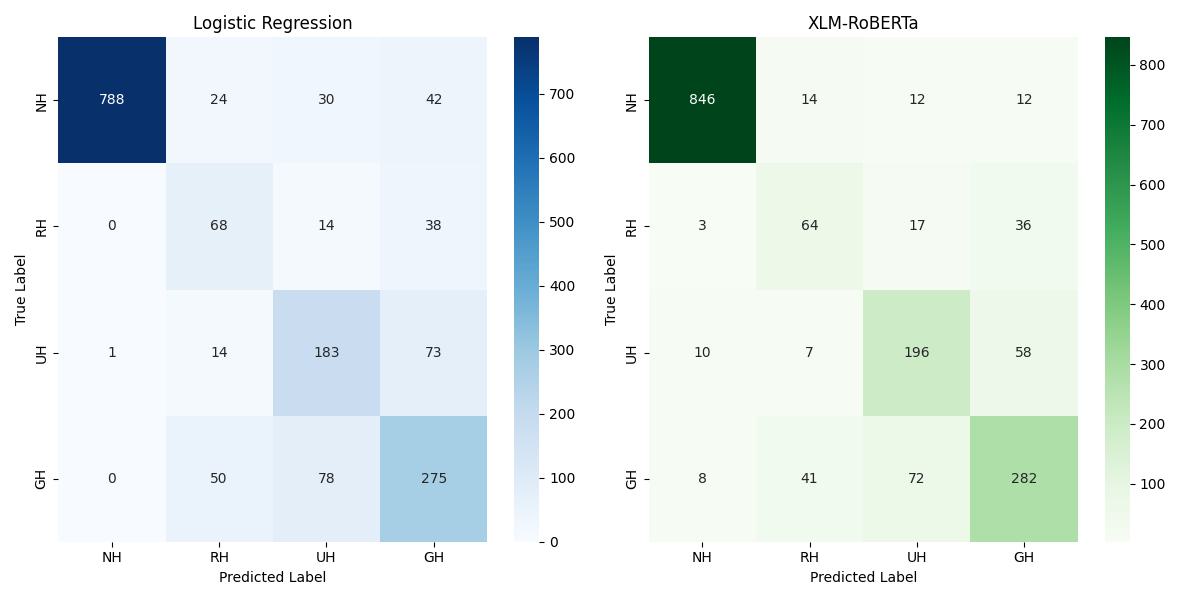}
    \caption{Urdu Dev Set Confusion Matrix}
    \label{fig:enter-label}
\end{figure}

\end{document}